\documentclass[conference]{IEEEtran}
\IEEEoverridecommandlockouts
\usepackage{amsmath,amssymb,amsfonts}
\usepackage{algorithmic}
\usepackage{graphicx}
\usepackage{textcomp}
\usepackage{xcolor}
\usepackage{multirow}
\usepackage{mdframed}
\usepackage{makecell}
\usepackage[
    natbib=true,
    style=numeric,
    sorting=none
]{biblatex}
\begin{document}

\title{LLM is Not All You Need: A Systematic Evaluation of ML vs. Foundation Models for text and image based Medical Classification
}

\author{\IEEEauthorblockN{1\textsuperscript{st} Meet Raval}
\IEEEauthorblockA{\textit{University of Southern California} \\
Los Angeles, USA \\
mkraval@alumni.usc.edu}
~\\
\and
\IEEEauthorblockN{2\textsuperscript{nd} Tejul Pandit*}
\IEEEauthorblockA{\textit{Palo Alto Networks} \\
Santa Clara, USA \\
tejulpandit96@gmail.com}
*Corresponding author
~\\
\and
\IEEEauthorblockN{3\textsuperscript{rd} Dhvani Upadhyay}
\IEEEauthorblockA{\textit{Dhirubhani Ambani University} \\
Gandhinagar, India \\
202001136@dau.ac.in}
}

\maketitle

\begin{abstract}
The combination of multimodal Vision-Language Models (VLMs) and Large Language Models (LLMs) opens up new possibilities for medical classification. This work offers a rigorous, unified benchmark by using four publicly available datasets covering text and image modalities (binary and multiclass complexity) that contrasts traditional Machine Learning (ML) with contemporary transformer-based techniques. We evaluated three model classes for each task: Classical ML (LR, LightGBM, ResNet-50), Prompt-Based LLMs/VLMs (Gemini 2.5), and Fine-Tuned PEFT Models (LoRA-adapted Gemma3 variants). All experiments used consistent data splits and aligned metrics. According to our results, traditional machine learning (ML) models set a high standard by consistently achieving the best overall performance across most medical categorization tasks. This was especially true for structured text-based datasets, where the classical models performed exceptionally well. In stark contrast, the LoRA-tuned Gemma variants consistently showed the worst performance across all text and image experiments, failing to generalize from the minimal fine-tuning provided. However, the zero-shot LLM/VLM pipelines (Gemini 2.5) had mixed results; they performed poorly on text-based tasks, but demonstrated competitive performance on the multiclass image task, matching the classical ResNet-50 baseline. These results demonstrate that in many medical categorization scenarios, established machine learning models continue to be the most reliable option. The experiment suggests that foundation models are not universally superior and that the effectiveness of Parameter-Efficient Fine-Tuning (PEFT) is highly dependent on the adaptation strategy, as minimal fine-tuning proved detrimental in this study.
\end{abstract}


\section{Introduction}

Advanced Artificial Intelligence (AI) is rapidly changing clinical procedures in the healthcare industry, especially in the areas of medical data processing[Multi-domain ABSA data generation] and diagnostic support. A paradigm change has been made possible by the development of Large Language Models (LLMs) and their multimodal extensions (VLMs), which allow scalable analysis of heterogeneous data, such as unstructured clinical text, medical pictures, and integrated reports. In the past, structured data, manual feature engineering, and well-known supervised machine-learning (ML) models like Logistic Regression (LR), Support Vector Machines (SVMs), or Gradient-Boosted Trees (GBTs) have been crucial for clinical classification tasks like diabetes prediction, mental-health disorder classification, or the detection of pathological findings in medical images. Although these classical methods offer robustness, computational efficiency, and crucial interpretability, they inherently struggle with the increasing volume of complex, high-dimensional, and unstructured or multimodal data generated in modern healthcare settings.

\subsection{The Rise of Foundation Models in Medicine}

Transformer-based LLMs, such as the Gemini \cite{geminiteam2025geminifamilyhighlycapable} and GPT \cite{brown2020languagemodelsfewshotlearners} families, have shown hitherto unprecedented capabilities in knowledge retrieval, complicated reasoning, natural language comprehension, and data generation tasks \cite{chen2024surveymedicalvisionandlanguageapplications, paperllmapplications, pandit2025multidomainabsaconversationdataset}. LLMs are now being used in RAG systems, which are showing remarkable efficiency in clinical natural language processing (NLP) for a variety of applications, including clinical text generation, clinical note categorization, patient-report summarization, decision support, and medical education\cite{heathcare_eval, Naved_Ravishankar_Colbert_Johnston_Slott_Luo_2025}. Simultaneously, Vision-Language Models (VLMs) \cite{bordes2024introductionvisionlanguagemodeling} have started to simplify medical imaging analysis by fusing strong image encoders with language-based outputs, making tasks like content-based image retrieval, automated report production, and diagnostic reasoning easier \cite{liu2024surveymedicallargelanguage}. These advances have the potential to significantly reduce the difficulty in implementing sophisticated categorization techniques in a wide range of medical inputs.

\subsection{Limitations in Current Benchmarking}

Despite this quick development and clear promise, there are still a number of serious flaws in the research that prevent these foundation models from being widely and intelligently adopted in high-stakes clinical settings:

\begin{itemize}
    \item Absence of Cross-Modality Alignment: Previous studies have mostly focused on text-based or image-based classification separately. Unified experimental frameworks that perform simultaneous and controlled comparisons across both modalities (text and image) utilizing the same assessment metrics and methodologies are critically lacking.
    \item Inconsistent Benchmarking Rigor: Evaluations frequently concentrate primarily on the superiority of one big model over another or compare only particular LLM paradigms (such as zero-shot prompting). The entire range of modeling techniques, classical ML pipelines, zero/few-shot prompt-based LLMs, and Parameter-Efficient Fine-Tuned (PEFT) LLMs \cite{han2024parameterefficientfinetuninglargemodels}, are rarely thoroughly benchmarked side by side under similar dataset splits and pre-processing conditions \cite{RyuKangChuYang2025}.
    \item Underexplored Complexity: Compared to more straightforward binary tasks, multiclass classification—which is crucial for differential diagnosis, particularly in medical imaging with VLMs and refined multimodal models—remains much understudied.
\end{itemize}

\subsection{Research Objective}

This work offers a thorough, four-task experimental comparison intended to provide empirical clarity on the performance landscape of contemporary foundation models vs well-established machine learning techniques in medical categorization in order to fill in these essential gaps. Four freely accessible datasets that cover binary and multiclass tasks in text and image modalities are used.

We evaluated classical ML pipelines (logistic regression, LightGBM), prompt-based LLMs (Gemini 2.5 Flash for text, Gemini 2.5 multimodal for images \cite{comanici2025gemini25pushingfrontier}), and fine-tuned LoRA-adapted \cite{hu2021loralowrankadaptationlarge} LLMs (Gemma3\_Instruct\_1B for text, Gemma3 + LoRA for vision \cite{gemmateam2025gemma3technicalreport}).

Our research aims to answer the following key questions:

\begin{itemize}
    \item \textbf{RQ1:} For binary and multiclass medical text classification problems, how do prompt-based and refined LLMs compare to traditional machine learning techniques?
    \item \textbf{RQ2:} How do Vision-LLM pipelines and classical CNNs fare in medical image categorization, both in zero-shot and fine-tuned settings?
    \item \textbf{RQ3:} When compared to standard ML in the medical field, under what particular circumstances (modality type, label complexity, dataset characteristics) do LLM-based techniques offer proven performance gains, or fail to do so?
\end{itemize}

Our study offers a robust, apples-to-apples comparison by rigorously harmonizing pre-processing, data splitting techniques, and evaluation criteria across all workloads. For medical AI practitioners, choosing between tried-and-true ML techniques and the rapidly changing field of big transformer-based solutions for practical healthcare deployment, the ensuing insights are crucial. 

The paper is structured such that Section \ref{relatedWork} offers Related Work, which reviews earlier studies on large-language models, vision-language models, and classical machine learning in the medical field. The datasets utilized are described in Section \ref{dataset}, together with information on classification complexity (binary vs. multiclass) and modalities (text vs. image). Our methodology, including the specific design of every pipeline and model, is described in Section \ref{method}. The experimentation results, metrics used, and performance outcomes are reported in Section \ref{exp}. Section \ref{conclusion} concludes with a discussion of future research directions.

\section{Related Work} \label{relatedWork}

The investigation of machine-learning (ML) and Large Language Model (LLM) methods for medical classification spans three primary research streams: (A) conventional ML for structured and unstructured medical data; (B) LLMs for medical text classification; and (C) multimodal and Vision-LLM (VLM) modelling in medical imaging. We reviewed each area and then outline how our work addresses the remaining critical gaps.

\subsection{Conventional ML for Medical Classification}

Historically, many medical diagnostic classification tasks (e.g., diabetes prediction, skin-lesion classification) have relied on well-understood feature engineering and classical supervised ML pipelines, such as Logistic Regression (LR), Support Vector Machines (SVMs), and Gradient-Boosted Trees (GBTs). These approaches benefit from simplicity, interpretability, relatively low computational cost, and established validation protocols. For example, ML pipelines that use TF-IDF and structured features have shown robust performance in clinical-note classification settings \cite{ZhouXuZhangXuGuoZhanFangDingWangXu}.

These traditional methods establish a necessary baseline that many newer works compare against, but they often struggle when the input is highly unstructured (free-text notes) or multimodal (image + text) and requires extensive, hand-designed feature extraction.

\subsection{LLMs for Medical Text Classification}

A growing body of literature investigates the use of LLMs for the classification of medical text and patient-reported narratives. Reviews such as \cite{ZhouXuZhangXuGuoZhanFangDingWangXu} and \cite{liu2024surveymedicallargelanguage} provide broad overviews of Medical-domain LLMs (Med-LLMs), discussing their architectures, applications, and trustworthiness, while emphasizing issues around domain adaptation and interpretability.

Systematic studies focusing on healthcare text classification with LLMs find that while zero-shot/few-shot prompting can yield promising results, domain-specific fine-tuning is often required to consistently match or exceed the performance of strong classical ML baselines \cite{sakib2025chatcheckuplargelanguage}. A large-scale empirical evaluation of zero-shot GPT-4 on patient self-reported searches, for instance, showed performance statistically similar to supervised NLP models deployed in clinical settings \cite{Naved_Ravishankar_Colbert_Johnston_Slott_Luo_2025}. Additional works explore multi-label classification via knowledge distillation of LLMs for lightweight deployment \cite{sakai2025kdhmltcknowledgedistillationhealthcare}. These studies collectively point to the opportunities and limitations of LLM-based methods: strong representational capacity but high sensitivity to domain adaptation, calibration, and data distribution shifts.

\subsection{Multimodal and Vision-LLM Models in Medical Imaging}

In medical imaging, large foundational models and Vision-Language Models (VLMs) are increasingly applied for tasks like classification, segmentation, report generation, and visual question answering. Surveys examining vision-language foundation models in medical imaging emphasize this growing area \cite{chen2024surveymedicalvisionandlanguageapplications, ZhouXuZhangXuGuoZhanFangDingWangXu, RyuKangChuYang2025}.

An example is the work demonstrating zero-shot and few-shot classification of biomedical images (e.g., MRI, chest X-ray) using domain-adapted VLMs \cite{van2024largevisuallanguagemodels}. Other research shows that when the training data is scarce, contrastive vision-language pre-training enables improved retrieval and classification in medical domains \cite{windsor2023visionlanguagemodellingradiologicalimaging}. Although the promise of VLM is high, these studies also highlight persistent challenges related to negation misunderstanding \cite{Adam}, interpretability, and the necessity of domain-specific adaptation \cite{heathcare_eval}. To date, however, few works benchmark VLMs directly against controlled classical CNN baselines in a unified, multi-task experimental framework.

\subsection{Critical Gaps \& Our Contribution}

Despite the breadth of the current literature, three critical deficiencies persist, which our work is specifically designed to resolve.

\begin{itemize}
    \item \textbf{Modality Isolation}: Existing studies overwhelmingly focus on single-modality evaluations (text-only or image-only), neglecting the need for an aligned, cross-modality comparison between diagnostic inputs.
    \item \textbf{Benchmarking Rigor}: A significant gap exists in the rigorous, controlled benchmarking of contemporary LLM and VLM pipelines against strong, established classical ML baselines under a unified and consistent evaluation protocol (i.e. identical data splits, metrics, and hyper-parameter constraints).
    \item \textbf{Multiclass VLM in Medicine}: The specific challenge of multiclass classification in medical imaging using Vision-LLMs remains substantially underexplored, limiting insights into model failure modes on complex, imbalanced tasks.
\end{itemize}

We uniquely contribute by implementing a four-task, cross-modal empirical framework (text-binary, text-multiclass, image-binary, image-multiclass). By systematically deploying classical ML, zero-shot/prompting LLM, and parameter-efficient fine-tuned (LoRA) LLM pipelines across both text and vision modalities, we establish the most comprehensive, apples-to-apples comparison to date. This rigorous design provides granular, evidence-based insights into the precise conditions under which foundation models deliver tangible, reliable performance gains over established ML techniques in the critical domain of medical classification.

\section{Dataset} \label{dataset}

Four publicly accessible medical datasets covering two modalities (text and picture), and two task settings (binary vs. multiclass) are chosen. As a result, under heterogeneous input representations, LLM-based pipelines can be systematically compared to traditional machine learning.

\subsection{Diabetes (Binary Text Classification)}

This dataset originates from the National Institute of Diabetes and Digestive and Kidney Diseases and is available on Kaggle \cite{Akturk_2020}. It contains 2,768 patient records with eight structured clinical indicators (e.g., 'Pregnancies', 'Glucose', 'BMI', 'Age') and a binary target variable 'Outcome' (0 = no diabetes, 1 = diabetes). The dataset was split into a training set of 2,214 records and a test set of 554 records. For LLM-based assessment, each feature row is translated into a natural language description , while traditional ML baselines consume the original structured numerical representation.

\subsection{Mental Health Disorders (Multiclass Text Classification)}

We use the ``Depression \& Mental Illness Classification Dataset" from Mendeley Data \cite{Mondol_2025}. This dataset includes 1,998 anonymized survey responses annotated into 12 distinct categories of mental health disorders (the target variable 'Depression\_Type'). The dataset was split into a training set of 1,598 records and a test set of 400 records. Similar to the diabetes task, entries are formatted into text prompts for LLM inference , while traditional ML pipelines use the structured and one-hot encoded features.

\subsection{Skin Cancer (Binary Image Classification)}

For the binary image task, we use the "Segmented Images of the Skin Cancer Dataset" from Kaggle \cite{Mhedhbi_2020}. The dataset contains 2,637 dermoscopic images divided into two classes: benign and malignant. The data was split as follows:

\begin{itemize}
    \item \textbf{Training Set:} 1,687 images
    \item \textbf{Validation Set:} 422 images
    \item \textbf{Test Set:} 528 images
\end{itemize}

All images were resized to 128x128 pixels with 3 color channels (RGB) and normalized.

\subsection{Respiratory Disease (Multiclass Image Classification)}

For multiclass image classification, we adopt the "Single-label Classification of Lung Diseases" dataset from Roboflow Universe \cite{single-label-classification-of-lung-diseases_dataset}. This dataset contains chest imaging samples pre-split into train, validation, and test sets, covering five categories: bacterial\_pneumonia, coronavirus, tuberculosis, viral\_pneumonia, and normal\_lung.

\begin{itemize}
    \item \textbf{Training Set:} 560 images (112 per class)
    \item \textbf{Validation Set:} 80 images (16 per class)
    \item \textbf{Test Set:} 160 images (32 per class)
\end{itemize}

All images were resized to 224x224 pixels with 3 color channels (RGB) and normalized.

\subsection{Summary}

Table \ref{tab:summary} provides a detailed summary of the tasks performed for the various medical datasets.

\begin{table}[htbp]
\caption{Summary of Datasets Used in This Study} \label{tab:summary}
\centering
\renewcommand{\arraystretch}{1.2} %
\begin{tabular}{|c|c|c|c|}
\hline
\textbf{\makecell{Dataset}} & \textbf{\makecell{Modality}} & \textbf{\makecell{Task\\Type}} & \textbf{\makecell{Target\\Labels}} \\
\hline
\makecell{Diabetes} & \makecell{Text} & \makecell{Binary} & \makecell{diabetic /\\non-diabetic} \\
\hline
\makecell{Mental Health\\Disorders} & \makecell{Text} & \makecell{Multiclass} & \makecell{multiple stressor-based\\categories} \\
\hline
\makecell{Skin Cancer\\(Segmented)} & \makecell{Image} & \makecell{Binary} & \makecell{malignant /\\benign} \\
\hline
\makecell{Respiratory\\Disease} & \makecell{Image} & \makecell{Multiclass} & \makecell{multiple respiratory\\conditions} \\
\hline
\end{tabular}
\label{tab_dataset_summary}
\end{table}

\section{Methodology} \label{method}

In this study, we design parallel experimental pipelines for text and image classification across four distinct medical datasets. The methodology enables a controlled comparison across modality (text/image), label complexity (binary/multiclass), and model class (traditional ML vs. LLM-based). All models were evaluated on a held-out test set using consistent data splits.

\subsection{Data Quality \& Pre-processing}

Before training any machine learning models for a text-based model, we performed pre-processing and data quality tests. 

\begin{figure}[htbp]
  \centering
  \includegraphics[width=0.45\textwidth]{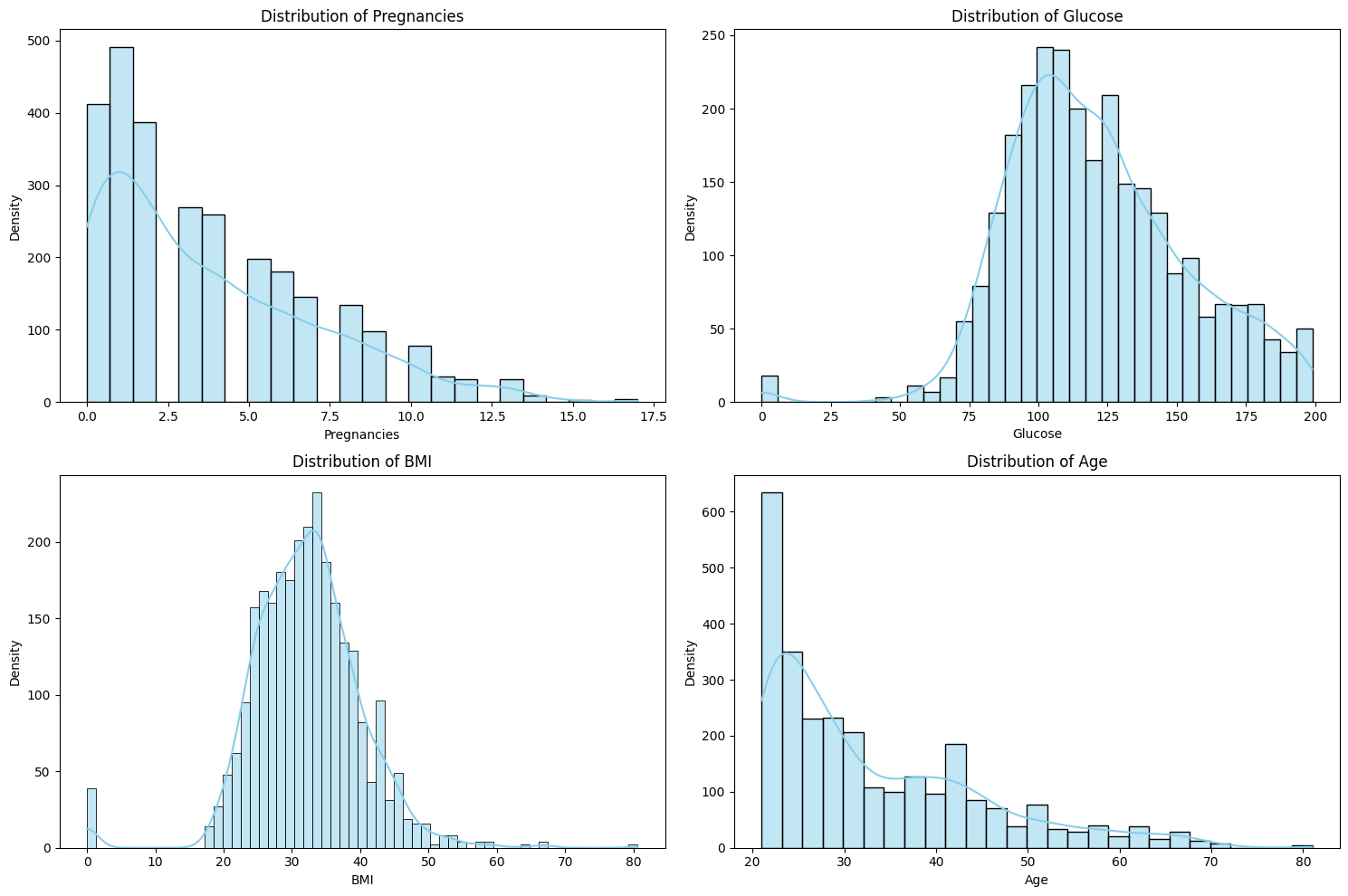}
  \caption{Numerical Feature Distributions for Diabetes Predictors}
  \label{img:hist}
\end{figure}

\begin{figure}[htbp]
  \centering
  \includegraphics[width=0.45\textwidth]{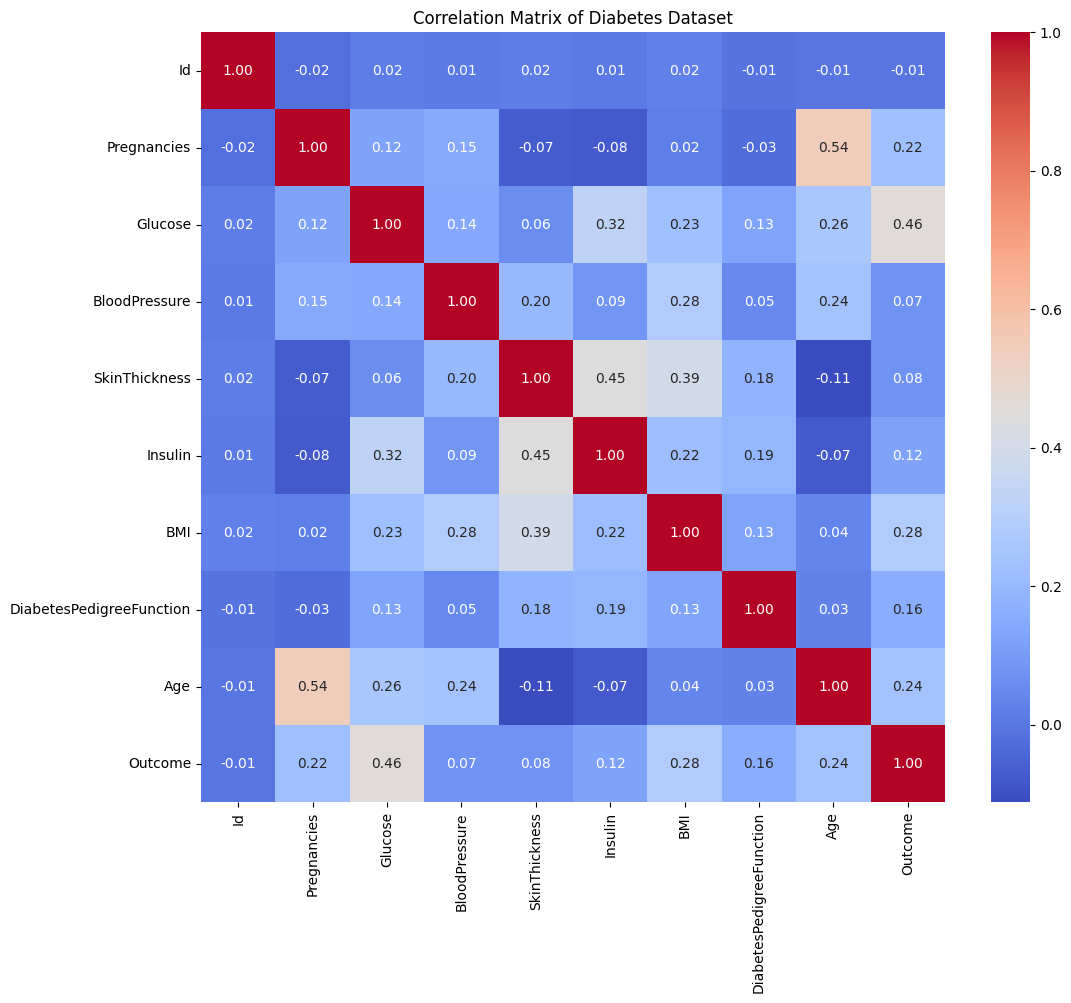}
  \caption{Correlation Matrix Heatmap for Diabetes Predictors}
  \label{img:corrmatrix}
\end{figure}

\begin{figure}[htbp]
  \centering
  \includegraphics[width=0.45\textwidth]{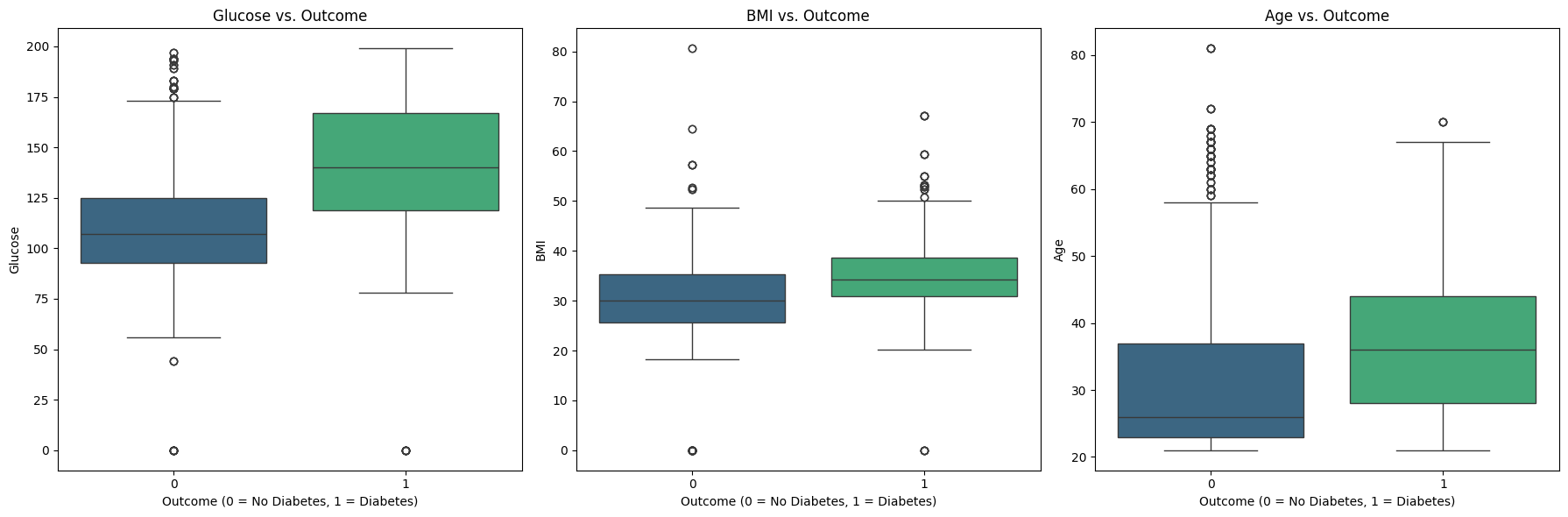}
  \caption{Relationship between Key Features and Outcome for Diabetes Predictors}
  \label{img:boxplots}
\end{figure}

For the Diabetes dataset, initial completeness was confirmed by the absence of missing values across all features. However, descriptive statistics showed that several medically significant attributes (such as \textit{Glucose}, \textit{BloodPressure}, \textit{SkinThickness}, \textit{Insulin}, and \textit{BMI}) contained a high proportion of zero values, which are physiologically implausible and likely represent placeholders or missing entries rather than valid measurements. Histograms and kernel density plots were used to examine the distributions of key predictors (such as \textit{Pregnancies}, \textit{Glucose}, \textit{BMI}, and \textit{Age}) (Fig.~\ref{img:hist}), revealing distinct ranges, skewness, and central tendency differences conditioned on diabetes status (Outcome $\in \{0,1\}$). A correlation matrix heatmap (Fig.~\ref{img:corrmatrix}) further demonstrated that higher \textit{Glucose}, \textit{BMI}, and \textit{Age} values are positively associated with the presence of diabetes. In addition, box plots contrasting Outcome = 0 vs. 1 (Fig.~\ref{img:boxplots}) highlighted consistently higher median values for these same features among diabetic subjects, indicating their importance for downstream classification.

To address the identified data-quality issues, biologically implausible zeros in ‘Glucose’, ‘BloodPressure’, ‘SkinThickness’, ‘Insulin’ and ‘BMI’ were replaced with the median of the non-zero values of each respective feature. Following cleaning, the dataset was stratified into training (2,214 samples) and testing (554 samples) subsets while preserving the original class distribution of the target variable.

For the Mental-Health dataset, initial validation similarly confirmed that there were no missing values across the feature space. However, the distribution of the target variable (\textit{Depression\_Type}) was highly imbalanced: the most common class (Type 9) contained 627 instances, whereas the least common class (Type 8) contained only 21 instances. This imbalance has direct implications for model evaluation and must be accounted for when reporting metrics.

\begin{figure}[htbp]
\centering
\includegraphics[width=0.45\textwidth]{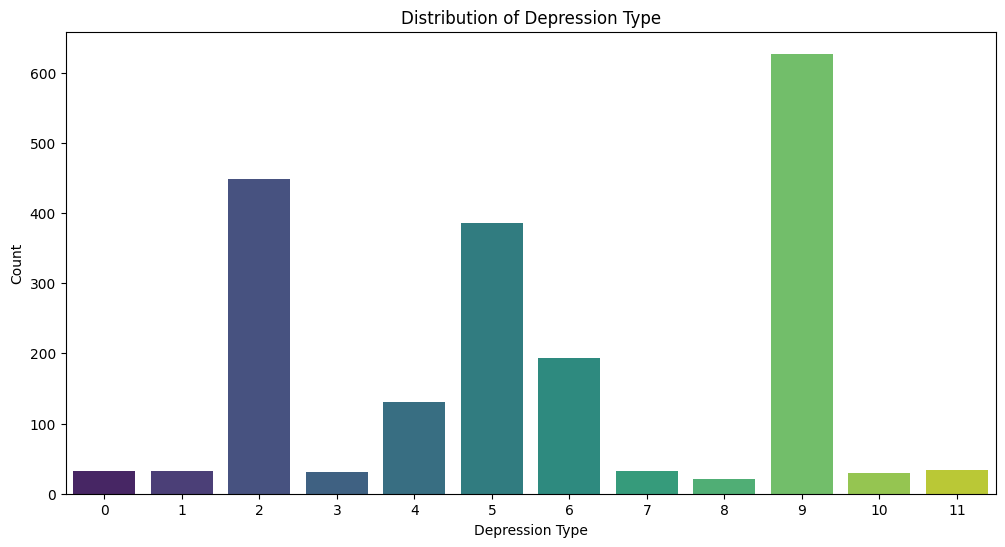}
\caption{Distribution of \textit{Depression Type}}
\label{img:mh_type_dist}
\end{figure}

\begin{figure}[htbp]
\centering
\includegraphics[width=0.45\textwidth]{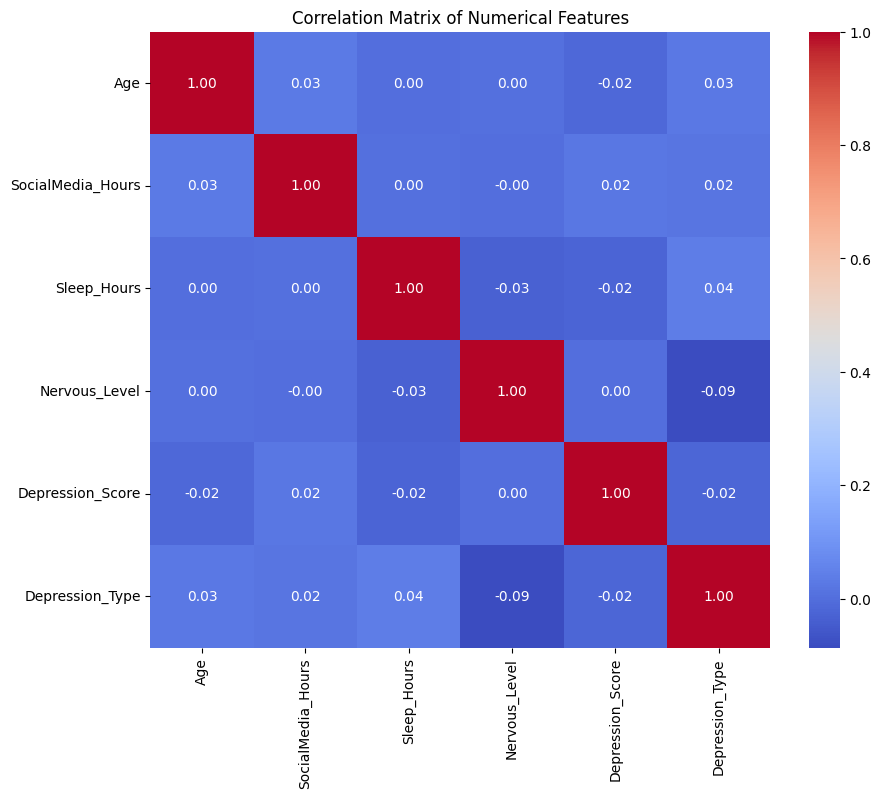}
\caption{Correlation Matrix for Numerical Features (Mental Health Dataset)}
\label{img:mh_corr}
\end{figure}

\vspace{1mm}
\noindent\textbf{Exploratory Inspectors -- Key Findings}
\begin{itemize}
    \item No missing values detected.
    \item The target variable \textit{Depression\_Type} is highly skewed across 10 classes as highlighted in Fig.-\ref{img:mh_type_dist}.
    \item Numerical features (\textit{Age}, \textit{SocialMedia\_Hours}, \textit{Sleep\_Hours}, \textit{Nervous\_Level}, \textit{Depression\_Score}) exhibit heterogeneous ranges and distribution shapes.
    \item Categorical features (\textit{Gender}, \textit{Education\_Level}, \textit{Employment\_Status}, \textit{Symptoms}) also demonstrate heterogeneous frequency distributions.
    \item Correlation analysis among numerical features provides additional insight into linear relationships, including correlations with the target as shown in Fig.-\ref{img:mh_corr}.
\end{itemize}

The severe class imbalance in \textit{Depression Type} was mitigated at training-time using class-weighted loss and class-weighted performance metrics (i.e. - weighted F1).

\subsection{Conventional Machine Learning Pipelines} \label{convml}

For the text-based tasks, we adopted two conventional baselines as described in the paper: a regularized Logistic Regression (LR) and a LightGBM (LGBM) classifier.

\begin{itemize}
    \item \textbf{Diabetes (Binary)}: The 8 numerical features were scaled using StandardScaler before being fed into the LR and LGBM models.
    \item \textbf{Mental Health (Multiclass)}: Numerical features were scaled using StandardScaler, and categorical features were one-hot encoded. The models were trained on this fully preprocessed dataset.
\end{itemize}

For the image-based tasks, we used standard deep learning baselines:

\begin{itemize}
    \item \textbf{Skin Cancer (Binary)}: A custom Convolutional Neural Network (CNN) was built with three convolutional/max-pooling blocks and a final dense layer for classification.
    \item \textbf{Respiratory Disease (Multiclass)}: A pre-trained ResNet-50 architecture was used, with its base layers frozen and a new classification head trained on the lung disease data.
\end{itemize}

\subsection{LLM-based Classification}

For the same tasks, we employed two LLM-based approaches: zero-shot prompting and parameter-efficient fine-tuning.

\subsubsection{Prompt-based Inference}

\begin{itemize}
    \item We used the Gemini 2.5 Flash model for all zero-shot tasks. To ensure reproducibility and transparency regarding the prompt engineering strategies employed, we detail the specific system instructions and user templates in Appendix \ref{sec:appendix_derivations}.
    \item For the Diabetes task, a prompt was engineered by instructing the model to act as a "medical data analyst" and return a single probability, which was then thresholded at 0.5 for classification.
    \item For the Mental Health task, an n-shot prompt was used, providing one example for each of the 12 depression types, and instructing the model to return only the predicted integer class.
    \item For the image tasks, specialized prompts were used to request classification of the provided image (e.g. "Benign" or "Malignant" for skin cancer).
\end{itemize}

\subsubsection{Fine-tuned LoRA Model}

\begin{itemize}
    \item We performed Parameter-Efficient Fine-Tuning (PEFT) using Low-Rank Adaptation (LoRA) on Gemma variants.
    \item For the text tasks (Diabetes and Mental Health), a Gemma3\_Instruct\_1B model was fine-tuned using Keras-hub. The models were trained for 1 epoch with a LoRA rank of 4, using the same prompt structures as the zero-shot Gemini models.
    \item For the image tasks, a Gemma 3 4B-PT vision variant was fine-tuned for 1 epoch using LoRA (rank $r=4$).
\end{itemize}

\subsection{Evaluation Protocol}

All models were evaluated on the held-out test set using consistent splits. We report performance using Accuracy, F1-Score, and Area Under the ROC Curve (AUC-ROC). For multiclass tasks, weighted-average F1-scores are used.

\section{Experiments} \label{exp}

We conducted four distinct experiments to benchmark the three model classes (Classical ML, Zero-Shot LLM, Fine-Tuned LLM) across the two modalities and two task complexities.

\subsection{Text Classification Tasks}

For both text-based tasks, the classical machine learning models (LightGBM and Logistic Regression) significantly outperformed both the zero-shot and fine-tuned LLM approaches. The performance of the LLMs on the structured data, even when formatted as text, was notably poor.

\begin{itemize}
    \item \textbf{Diabetes (Binary)}: The LightGBM model achieved near-perfect classification (0.9982 Accuracy), demonstrating its effectiveness for this structured dataset. The fine-tuned Gemma model (0.6661 Accuracy) performed better than the zero-shot Gemini model (0.4224 Accuracy) but was still far surpassed by the classical models.
    \item \textbf{Mental Health (Multiclass)}: Both LightGBM and Logistic Regression achieved close to perfect accuracy on the test set, suggesting that the task may be linearly separable or easily solvable with the one-hot encoded features. In contrast, the zero-shot Gemini model struggled with the 12-class problem (0.2807 Accuracy) , and the LoRA-tuned Gemma model failed completely, achieving an accuracy of only 0.0150.
\end{itemize}

The metrics in Table \ref{tab:text_stats} represents the results for test-set for text classification tasks using the 4 different models.

\begin{table}[htbp]
\caption{Text Classification Test Set Results} \label{tab:text_stats}
\centering
\renewcommand{\arraystretch}{1.2} %
\begin{tabular}{|c|c|c|c|c|}
\hline
\textbf{\makecell{Task}} & 
\textbf{\makecell{Model}} & 
\textbf{\makecell{Accuracy}} & 
\textbf{\makecell{F1-Score}} & 
\textbf{\makecell{ROC\\AUC}} \\
\hline
\multirow{4}{*}{\makecell{Diabetes\\(Binary)}} 
 & \makecell{LightGBM\\(ML)} & \textbf{0.9982} & \textbf{0.9974} & \textbf{0.9992} \\
 & \makecell{Logistic\\Regression\\(ML)} & 0.7726 & 0.6272 & 0.8355 \\
 & \makecell{Gemini 2.5\\(Zero-Shot)} & 0.4224 & 0.4576 & 0.4898 \\
 & \makecell{Gemma 1B\\(LoRA)} & 0.6661 & 0.0887 & 0.5194 \\
\hline
\multirow{4}{*}{\makecell{Mental Health\\(Multi)}} 
 & \makecell{LightGBM\\(ML)} & \textbf{0.9866} & \textbf{0.9821} & \textbf{0.9902} \\
 & \makecell{Logistic\\Regression\\(ML)} & 0.97 & 0.9655 & 0.9771 \\
 & \makecell{Gemini 2.5\\(Zero-Shot)} & 0.2807 & 0.2660 & N/A \\
 & \makecell{Gemma 1B\\(LoRA)} & 0.0150 & 0.0004 & N/A \\
\hline
\end{tabular}
\label{tab_text_classification_results}
\end{table}

\subsection{Image Classification Tasks}

For image-based tasks, the results were more varied. The classical CNN baseline remained the most robust choice for the binary task, while the zero-shot VLM was competitive in the multiclass setting.

\begin{itemize}
    \item \textbf{Skin Cancer (Binary)}: The custom-built CNN baseline achieved a strong accuracy of 0.8277. The zero-shot Gemini 2.5 Flash model performed significantly worse (0.5890 Accuracy). The LoRA-tuned Gemma 3 4B-PT model failed to learn effectively, resulting in an F1-score of 0.2012.
    \item \textbf{Respiratory Disease (Multiclass)}: The ResNet-50 baseline and the zero-shot Gemini 2.5 Flash model showed nearly identical performance (0.4812 vs 0.4875 Accuracy), suggesting that for certain complex multiclass vision tasks, a zero-shot VLM can be as effective as a trained CNN baseline. The LoRA-tuned Gemma model again performed the worst (0.3688 Accuracy).
\end{itemize}

Table \ref{tab:img_stats} represent the test set results for image classification tasks using the 3 different approaches.

\begin{table}[htbp]
\caption{Image Classification Test Set Results} \label{tab:img_stats}
\centering
\renewcommand{\arraystretch}{1.2} %
\begin{tabular}{|c|c|c|c|}
\hline
\textbf{\makecell{Task}} & 
\textbf{\makecell{Model}} & 
\textbf{\makecell{Accuracy}} & 
\textbf{\makecell{F1-Score}} \\
\hline
\multirow{3}{*}{\makecell{Skin Cancer\\(Binary)}} 
 & \makecell{CNN\\(ML Baseline)} & \textbf{0.8277} & \textbf{0.8154} \\
 & \makecell{Gemini 2.5\\(Zero-Shot)} & 0.5890 & 0.3439 \\
 & \makecell{Gemma 3 4B-PT\\(LoRA)} & 0.5455 & 0.2012 \\
\hline
\multirow{3}{*}{\makecell{Respiratory\\(Multi)}} 
 & \makecell{ResNet-50\\(ML Baseline)} & 0.4812 & \textbf{0.4663} \\
 & \makecell{Gemini 2.5\\(Zero-Shot)} & \textbf{0.4875} & 0.4638 \\
 & \makecell{Gemma 3 4B-PT\\(LoRA)} & 0.3688 & 0.2127 \\
\hline
\end{tabular}
\label{tab_image_classification_results}
\end{table}

\subsection{Summary of Findings}

The experiments confirm the paper's central thesis. Across all four tasks, traditional machine learning models set a high standard that was consistently unmet by contemporary transformer-based techniques.

\begin{itemize}
    \item On the \textbf{structured text-based datasets}, the classical ML models (LGBM, LR) were vastly superior, achieving perfect or near-perfect scores. The LLM approaches, both zero-shot and fine-tuned, performed very poorly in comparison.
    \item On the \textbf{image-based datasets}, a classical CNN baseline provided the best performance for the binary task. The zero-shot VLM (Gemini 2.5) showed competitive performance on the multiclass task, but was still not superior.
    \item The \textbf{LoRA-adapted Gemma} variants continuously showed the worst performance across all four tasks, indicating that minimal fine-tuning (1 epoch) was insufficient for domain-specific adaptation.
\end{itemize}

These results demonstrate that for these medical categorization scenarios, established machine learning models remain the most dependable option.

\subsection{Sensitivity Analysis on Adaptation Strategy}

To address the poor performance of the initial fine-tuning experiments, we conducted a sensitivity analysis on the Diabetes dataset to evaluate the impact of LoRA rank and training duration.

\subsubsection{Impact of LoRA Rank}
We hypothesized that the rank ($r=4$) might be insufficient for capturing feature interactions. However, increasing the LoRA rank to $r=8$ resulted in a performance degradation compared to the baseline configuration:

\begin{itemize}
    \item \textbf{Accuracy}: 0.6606
    \item \textbf{F1 Score}: 0.0693
    \item \textbf{ROC AUC Curve}: 0.5128
\end{itemize}

This suggests that simply increasing the number of trainable parameters does not inherently solve the domain adaptation challenge and may lead to overfitting on small datasets.

\subsubsection{Impact of Training Duration (Multi-Epoch)}
In contrast to rank adjustments, extending the training duration proved highly effective. While the 1-epoch baseline yielded poor generalization, extending the training to 5 and 10 epochs resulted in substantial gains, as shown in Table \ref{tab:sensitivity_training_duration}.

\begin{table}[!t]
\centering
\caption{Sensitivity Analysis: Training Duration (Diabetes)}
\label{tab:sensitivity_training_duration}
\small
\begin{tabular}{|c|c|c|c|}
\hline
\textbf{Epochs} & \textbf{Accuracy} & \textbf{F1 Score} & \textbf{ROC AUC} \\
\hline
5  & 0.8141 & 0.7049 & 0.7738 \\
10 & 0.8347 & 0.7152 & 0.7987 \\
\hline
\end{tabular}
\end{table}

These findings indicate that while PEFT is capable of learning the task, it requires significantly more epochs than typically advertised for "few-shot" or "efficient" adaptation to compete with classical baselines.

\section{Conclusion} \label{conclusion}

Even if classical machine learning models perform almost flawlessly on text-based tasks (LightGBM, for example, achieves 0.9982 accuracy on the Diabetes dataset), which may indicate easy separability or possible data leaking, this merely highlights Foundation Models' flaws. On a task that Logistic Regression completed flawlessly, a zero-shot LLM ($\sim$0.42 accuracy) and a 1-epoch fine-tuned LLM ($\sim$0.66) failed. The findings revealed a critical reliability gap and showed that LLMs struggle to interpret structured tabular data that is simple for standard statistical methods, requiring extensive, resource-intensive adaptation ($\ge$10 epochs). This discrepancy has significant implications for clinical deployment: LLMs showed greater volatility even under constrained decoding, whereas classical models provide deterministic, well-calibrated probabilities necessary for risk scoring; additionally, the latency and carbon cost of VLM-based inference for binary diagnosis are orders of magnitude higher than those of Logistic Regression or ResNet inference, without corresponding accuracy gains; and LLMs introduce additional safety risks like hallucination and format non-compliance (partially mitigated here through strict prompting). Classical models consistently proved the most dependable and efficient across our four-task medical benchmark, achieving near-perfect performance on structured text tasks when Foundation Models were unreliable. Our PEFT technique failed in every experiment, even though the zero-shot Gemini VLM was remarkably competitive with a trained ResNet-50 on the multiclass picture task: Limited fine-tuning can be detrimental, as demonstrated by the poor performance of LoRA-adapted Gemma models. The sensitivity analysis demonstrates that prolonged training (10 epochs) can reduce, but not eliminate, the performance gap. These findings empirically demonstrate that Foundation Models cannot fully replace traditional methods, as their efficacy depends on data-intensive, well-designed adaptation pipelines. Overall, for many common medical classification tasks, ``LLM is not all that you need.”

\printbibliography

\appendix %
\section{Appendix: Prompt Templates} %
\label{sec:appendix_derivations}
\subsection{Diabetes}
\begin{mdframed}[linewidth=0.5pt, leftmargin=0pt, rightmargin=0pt, innertopmargin=2pt, innerbottommargin=2pt]

\textit{user\_prompt\_template} = Based on the following patient data, what is the probability that the 'Outcome' is 1 (Positive for Diabetes)?

Patient Data:

- Pregnancies: \texttt{\{Pregnancies\}}

- Glucose: \texttt{\{Glucose\}}

- Blood Pressure: \texttt{\{BloodPressure\}}

- Skin Thickness: \texttt{\{SkinThickness\}}

- Insulin: \texttt{\{Insulin\}}

- BMI: \texttt{\{BMI\}}

- Diabetes Pedigree Function: \texttt{\{DiabetesPedigreeFunction\}}

- Age: \texttt{\{Age\}}

Your response must be a single floating-point number representing the probability (p), without any other text, explanation, or context. For example, a valid response would be: ``0.85".

\textit{system\_instruction} = You are an expert medical data analyst specializing in diabetes detection. Your task is to perform a binary classification based on a patient's health metrics. You will be provided with a set of 8 numerical features. Your job is to analyze these features and determine the probability that the patient has a positive 'Outcome' (Outcome=1), which indicates diabetes. The output must be a single numerical value between 0.0 and 1.0.
\end{mdframed}

\subsection{Mental Health}

\begin{mdframed}[linewidth=0.5pt, leftmargin=0pt, rightmargin=0pt, innertopmargin=2pt, innerbottommargin=2pt]

\textit{user\_prompt\_template\_multi} = USER PROMPT: Predict Depression Type

Analyze the following input data (clinical notes, features, historical context, etc.) and, strictly following the System Instruction, output the predicted `Depression\_Type'.

Based on the following patient data, classify the 'Depression\_Type'.

\textit{system\_instruction\_multi} = **Goal:** The model's sole task is to predict the 'Depression\_Type' based on the input data provided in the User Prompt.

**Strict Output Protocol (The Three Absolute Rules):**

1.  **Content Exclusivity:** The entire response **must** consist of only one element: the predicted 'Depression\_Type' value.

2.  **Format and Purity:** The value **must** be a **single, bare integer**. Do not include any surrounding characters, punctuation (periods, commas), labels, text, explanation, reasoning, quotes, spaces, or newlines.

3.  **Range Enforcement:** The predicted integer **must** be a positive digit from the closed set [0, 9].

**Examples of Non-Compliant Outputs (Strictly Forbidden):**

* ``The type is 7"

* Type: 5

* `9'

* 7.0

* [4]

**Example of a Valid, Compliant Output:**
3

\end{mdframed}

\subsubsection{Skin Cancer}

\begin{mdframed}[linewidth=0.5pt, leftmargin=0pt, rightmargin=0pt, innertopmargin=2pt, innerbottommargin=2pt]

\textit{user\_prompt\_template\_multi} = You are an AI dermatology assistant performing a visual analysis. Your task is to classify the skin lesion in the provided image.

Please follow these steps:

1.  **Analyze Image:** Examine the image for the ``ABCDE" criteria of melanoma:

    * **Asymmetry:** Is the lesion asymmetrical?
    
    * **Border:** Are the borders irregular, notched, or poorly defined?
    
    * **Color:** Is there variation in color (e.g., multiple shades of brown, black, red, or blue)?
    
    * **Diameter:** Is the diameter larger than 6mm (approx. the size of a pencil eraser)?
    
    * **Evolving:** (You cannot determine this, but note any features suggesting change or evolution).

2.  **Provide Reasoning:** Based on your analysis of these features, provide a step-by-step reasoning.

3.  **Final Classification:** Conclude with your final classification: **Benign** or **Malignant**.

**Output Format:**

* **Analysis:** [Your findings for Asymmetry, Border, Color, Diameter]

* **Reasoning:** [Your explanation linking the findings to the classification]

* **Classification:** [Benign/Malignant]

\end{mdframed}

\subsection{Respiratory Disease}

\begin{mdframed}[linewidth=0.5pt, leftmargin=0pt, rightmargin=0pt, innertopmargin=2pt, innerbottommargin=2pt]

\textit{user\_prompt\_template\_multi} = Classify the lung disease in the provided image.
Output only one of the following class names: bacterial\_pneumonia, normal\_lung, coronavirus, viral\_pneumonia, tuberculosis.

\end{mdframed}
\end{document}